\RequirePackage{fix-cm}
\documentclass[twocolumn]{svjour3} 
\smartqed 
\usepackage{graphicx}
\usepackage{amsmath}
\usepackage{bbm}
\begin{document}

\title{ Comparison of Algorithms that Detect Drug Side Effects using Electronic Healthcare Databases
}

\titlerunning{Comparison of Algorithms that Detect Drug Side Effects} 

\author{Jenna Reps$^1$ \and
Jonathan M. Garibaldi$^1$ \and Uwe Aickelin$^1$ \and Daniele Soria$^1$ \and Jack Gibson$^2$ \and Richard Hubbard$^2$
}


\institute{1 \at
Intelligent Modelling \& Analysis, NG8 1BB \\
Tel.: ++44 (0) 115 95 14299\\
\email{psxjr1@nottingham.ac.uk} 
\and
2 \at
Clinical Sciences Building, Nottingham City Hospital, NG5 1PB
}

\date{Received: date / Accepted: date}

\maketitle

\begin{abstract}

The electronic healthcare databases are starting to become more readily available and are thought to have excellent potential for generating adverse drug reaction signals. The Health Improvement Network (THIN) database is an electronic healthcare database containing medical information on over 11 million patients that has excellent potential for detecting ADRs. In this paper we apply four existing electronic healthcare database signal detecting algorithms (MUTARA, HUNT, Temporal Pattern Discovery and modified ROR) on the THIN database for a selection of drugs from six chosen drug families. This is the first comparison of ADR signalling algorithms that includes MUTARA and HUNT and enabled us to set a benchmark for the adverse drug reaction signalling ability of the THIN database. The drugs were selectively chosen to enable a comparison with previous work and for variety. It was found that no algorithm was generally superior and the algorithms' natural thresholds act at variable stringencies. Furthermore, none of the algorithms perform well at detecting rare ADRs.


\keywords{Adverse drug event \and Electronic healthcare database \and Longitudinal observational database \and MUTARA \and HUNT \and Observed Expected Ratio \and disproportionality methods}
\end{abstract}

\section{Introduction}
\label{intro}

It is an unavoidable consequence that prescription drugs frequently cause unwanted side effects due to the unpredictability of how a drug will interact with all the body functions \cite{Story1974}. When a negative side effect has been associated with a drug, it is referred to as an adverse drug reaction (ADR). Clinical trials are used as a means to identify common ADRs, but are not suitable at detecting all the possible ADRs. The reason clinical trials cannot find all the ADRs is due to the trials being limited by time constraints, sample population size, sample population bias and unrealistic conditions \cite{Primohamed2004}. For example, outside of a clinical trial it is common practice for a patient to be taking multiple prescription drugs but not all co-prescriptions can be monitored during clinical trials. It is also common for certain subpopulations such as children or pregnant women to be underrepresented in trials due to ethical reasons. As a consequence, post marketing surveillance is constantly required to identify any previously undiscovered ADR throughout the time a drug is actively prescribed. Pharmacovigilance signal detection is the process of identifying potential adverse reactions to a drug that were previously unknown. Once a tentative signal is generated, it is further evaluated to confirm causation between the drug and adverse drug reaction, if causation is shown, we refer to the signal as a true signal, if causation is not shown, it is a false signal. 

The majority of the traditional posts marketing surveillance techniques make use of spontaneous reporting system (SRS) databases, databases containing voluntary records of suspected drug and ADR pairs. The SRS algorithms generate tentative ADR signals by finding medical events that occur disproportionally more often after a specific drug compared to any drug in the database. The existing SRS algorithms include calculating measures frequently used in epidemiology such as the reporting odds ratio \cite{Puijenbroek2002} and proportional reporting ratio \cite{evans2001}, a Bayesian approach \cite{Bate1998} and an Empirical Bayes approach \cite{DuMouchel1999}. Due to the voluntary nature of these databases information is often missing, incorrect or duplicated \cite{Almenoff2005}. It is also believed that some ADRs may not be detectable by mining SRS databases due to under-reporting \cite{Alvarez1998}, for example people may not be bothered to report less severe side effects or may not notice a very rare ADR. The existing algorithms applied to SRS databases have previously been described as filters rather than definitive ADR detectors as the tentative signals they generate still require further investigation to determine if they are true or false. 

Recently a new type of medical database, the Electronic Healthcare Database (EHD), has started to attract attention for scientific research and is expected to become a fundamental component in future pharmacovigilance \cite{Wilke2011}. It is common for this type of database to contain a wealth of information such as complete medical and drug prescription histories for a patient. These databases offer a different perspective for detecting ADRs than the SRS databases and current research includes developing methods to mine EHDs or integrating EHDs and other healthcare databases together and mining the combination \cite{Coloma2011} \cite{MiniSentinel2008}. 

There has been a recent focus on developing algorithms that can mine EHDs and a variety of algorithms exist. A range of different methodologies have been incorporated into signalling ADRs including case control approaches \cite{Rosenberg2005}, cohort approaches \cite{Ryan2009}, modifying older algorithms developed for SRS databases \cite{Curtis2008} \cite{Zorych2011}, developing sequential pattern mining based algorithms aiming to find dependencies between drugs and medical events by using a control group \cite{Jin2006} \cite{Jin2010}, investigating temporal changes \cite{Noren2010} or by calculating the log likelihood over time \cite{Brown2007}. One previous study applied and compared a range of ADR signalling algorithms including the SRS adapted disproportionality algorithms and the Temporal Pattern Discovery (TPD) algorithm to multiple EHDs \cite{Ryan2012}. They showed that the algorithms investigated tended to return many false positive associations and that there was no clear optimal algorithm, as the most suitable algorithm depends on the desired trade off between the sensitivity and specificity. Another study concluded that there were little difference between the signalling ability of the SRS adapted disproportionality algorithms and a selection of cohort and case control algorithms \cite{Schuemie2012b}. To date there has been no comparison that includes the ADR signalling algorithms known as Mining Unexpected Temporal Association Rules given the Antecedent (MUTARA) \cite{Jin2006} and Highlighting UTARs Negating TARs (HUNT) \cite{Jin2010}. Both MUTARA and HUNT offer a unique perspective for signalling ADRs for a specific drug of interest as they are case control methods that incorporate a patient level filter. It is of interest to investigate how MUTARA and HUNT compare with the SRS adapted disproportionality algorithms and the TPD algorithm as these have been extensively investigated. 


An EHD known as The Health Improvement Network (THIN) is a UK database that contains time stamped medical data on over 11 million patients and has excellent potential for ADR signal detection. Recent work has investigated the suitability of ADR signalling using the THIN database in it's raw form compared to a mapped form that enables it to be integrated with other EHDs \cite{Zhou2013}. The study showed that many THIN records were unable to be mapped but this did not appear to limit the signalling ability of the mapped THIN database. In this paper, we applied a selection of ADR signalling algorithms to the raw THIN database. This enabled a benchmark to be determined that can be used to aid the development of future ADR signalling algorithms that are specific to the THIN database and can utilise all the information contained within. This is important as we do not know all the possible ADRs for any drug so there is no `golden standard' to compare ADRs signalling algorithms against. Furthermore, this comparison enables us to determine the rare ADR signalling potential of each algorithm investigated as this has not been previously studied. This is important as algorithms mining the THIN database may be more likely to signal rare ADRs (occurring in less than 1 in 1000 patients) compared to algorithms mining the SRS databases due to the THIN database not relying on voluntary reporting. Although rare ADRs do not occur often, if they are severe (such as liver failure or death), it is important to identify them and algorithms capable of signalling rare but severe ADRs would greatly improve current healthcare. 

In this paper we compare four existing algorithms (HUNT,  MUTARA, TPD and modified ROR) by applying them to the THIN database (www.thin-uk.com) consisting of general practice records for patients registered at participating practices within the UK. In previous comparisons applied to other EHDs the algorithm's general ability of ranking ADR as well as it's ability of signalling ADRs at the natural threshold have been investigated. For consistency, these measures are also determined in this study on the THIN database. 

The modified ROR and the TPD algorithm both have natural signalling thresholds (the value of the lower confidence interval is greater than 1 or 0 respectively). HUNT and MUTARA were developed to return a ranked list of potential ADRs but MUTARA has a natural threshold for signalling ADRs when the unexpected-leverage is greater than 0. Unfortunately, HUNT does not have a natural threshold so in this paper, for the natural threshold comparison, we implement a threshold of signalling the top 10\% of the ranked events. The general ability of identifying ADRs, independent of their natural signalling thresholds, can be observed by using rank based measures such as the area under the Receiver Operator Characteristic (ROC) and the average precision. The rank based measures are calculated by considering the top $n$ ranked events by each algorithm to be the tentative signals generated by each algorithm. Although technically a signal corresponds to an unknown potential ADR, for evaluation purposes in this paper we generate signals by considering that there are no known ADRs (so signals corresponding to a drug and adverse reaction are generated even if the drug is currently known to cause the adverse reaction) and use the known ADR knowledge to determine the number of true signals by finding how many of the known ADRs are signalled by each algorithm. So, we use the number of known ADRs occurring in the top $n$ ranked events by each algorithm to calculate the number of true signals generated by each algorithm where $n$ determines the signalling threshold. The sensitivity and specificity can then be evaluated. 

The continuation of this paper is as follows, section 2 gives a background on the existing algorithms describing the databases they were previously applied to and previous results. Section 3 contains information about the THIN database, the drugs investigated and the method applied in this paper to compare the algorithms. Section 4 presents the results of the comparison method and is followed by a discussion in section 5. This paper finishes with the conclusion in section 6.

\section{Algorithms}
\label{sec:1}

The EHDs do not contain direct links between drugs and medical events that are potential ADRs but these links can be inferred by using the temporal information in the database and finding all the medical events that occur within a set time period after the drug is prescribed. In this paper we are investigating the ability to detect immediately occurring ADRs so we consider all medical events that occur within $30$ days of the drug prescription as possible drug and ADR pairs. Justification for investigating the $30$ day time period after a drug prescription is that this is a trade off between having a long enough time period that under-reporting will be minimised (as patients have time to report any side effects) while reducing the amount of noise caused by a long time period. After finding all the medical events that occur within $30$ days of the drug prescription, the pharmacovigilance algorithms developed for EHDs then calculate a dependency measure between the drug and each medical event and return a list of all medical events that occurred within $30$ days of the prescription for any patient ranked by the dependency measure. The medical event ranked 1 is the medical event that appears to have the greatest dependency on the drug of interest having been previously prescribed. The existing EHD algorithms calculate the dependency measure differently and the majority implement a filter. We described the existing algorithms and the dependency measures used in the sections below.

\subsection{SRS Algorithms}
\subsubsection{SRS Algorithms \& Background}
The SRS algorithms were originally developed for general SRS databases where the actual rate that a drug is prescribed and the rate that a medical event occurs is unknown, as SRS databases only contain data on the drug prescriptions that may have resulted in an ADR. Consequently, the SRS algorithms estimate the background rate that a medical event occurs by finding out how often the medical event is reported with any drug in the database. Medical events that are reported disproportionally more often with the drug of interest compared to all the other drugs in the database are then ranked highly as suspected ADRs. These algorithms make use of a contingency table, see Table \ref{tab:srs}, summarising the number of reports that contain (or do not contain) the drug and event of interest.
\begin{table}
\centering
\caption{Contingency table used in existing SRS methods.}
\label{tab:srs}
\begin{tabular}{c|cc} 
& Event j =Yes & Event j =No \\ \hline
Drug i=Yes & w$_{00}$ & w$_{01}$ \\
Drug i=No & w$_{10}$ & w$_{11}$ \\ 
\end{tabular}
\end{table}

One example of an SRS algorithm is the reporting odds ratio (ROR),
\begin{equation}
ROR= \frac{w_{00}/w_{10}}{w_{01}/w_{11}}
\end{equation}
where a measure of how often the medical event occurs after the drug being investigated relative to how often it occurs after any other drug ($\frac{w_{00}}{w_{10}}$) is divided by how often any other medical event occurs after the drug being investigated relative to how often they occur after any other drug ($\frac{w_{01}}{w_{11}}$).

\subsubsection{SRS algorithms Previous Results}
Existing work by Zorych {\it et al.} applied the `modified SRS algorithms' to a simulated EHD database and a real EHD database \cite{Zorych2011}. They first transformed the EHD databases into an SRS style database containing reports of potential drug and ADR pairs by using temporal relations, in a similar way as described at the beginning of this section, see Fig. \ref{fig:srstrans}. After transforming the EHD data into SRS data they then applied the existing SRS disproportionality algorithms. 

\subsubsection{Implementation of the modified SRS algorithms in this study} 
\begin{figure*}[t]
\centering
\includegraphics[width=0.7\textwidth]{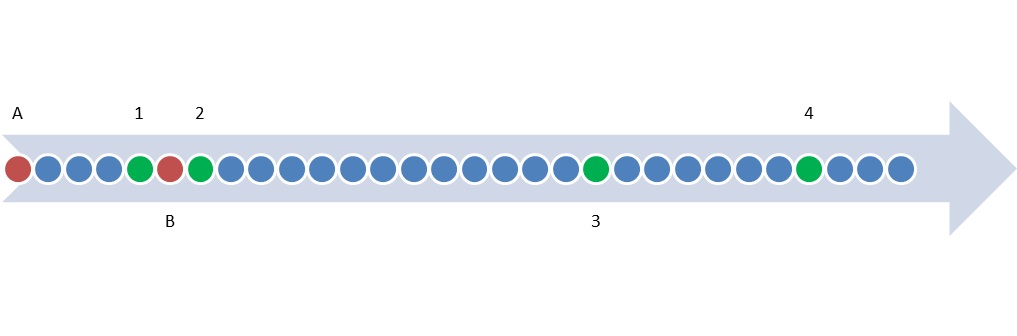}
\caption{ The line represents the THIN data for patient Pat1 with the arrow showing the direction of time and each circle is a day, with red circles representing a prescription of a drug and green circles representing an occurrence of a medical event. To transform this into SRS style data, the drugs are paired with all the medical events that occur within 30 days, so the THIN data represented in this Figure is transformed into seven SRS reports for the Pat1. The SRS style reports consist of four SRS data entries for drug A (Pat1-drug A -medical events 1,Pat1-drug A-medical event 2,Pat1-drug A- medical event 3 and Pat1-drug A-medical event 4) and three SRS data entries for Drug B (Pat1-Drug B-medical event 2, Pat1-Drug B-medical event 3 and Pat1-drug B-medical event 4).} 
\label{fig:srstrans}
\end{figure*}

In this paper we use the `Spontaneous reporting system' style transformation \cite{Zorych2011}, where SRS style reports consisting of a patient, drug prescription and possible ADR are inferred from the EHD by discovering all the medical events that occur within 30 days of a drug prescription. Fig. \ref{fig:srstrans} illustrates an example of the transformation.
After the THIN database is transformed, as described in Fig. \ref{fig:srstrans}, the ROR is applied. The left bound of the 90\% confidence interval of the ROR ($ROR_{05}$) is calculated for every event that occurs in at least three reports for the drug of interest ($w_{00} \geq 3$) and the medical events are ranked based on the $ROR_{05}$, see Eq. \ref{eq:ror} \cite{Puijenbroek2002} \cite{Zorych2011}. We chose to use the left bound of the 90\% confidence interval of the ROR as previous work showed that the $ROR_{05}$ was better at ranking medical events in terms of how likely they are to be ADRs than the ROR \cite{Zorych2011}. So the measure of dependency used to rank the medical events when applying the modified SRS algorithms to the THIN database is the chosen SRS algorithm value (in this case the $ROR_{05}$) calculated on the SRS style transformed database. 

\begin{equation}
\label{eq:ror}
\begin{split}
ROR_{05} = exp ( &ln(\frac{w_{00}/w_{10}}{w_{01}/w_{11}}) \\
&-1.645*\sqrt{ \frac{1}{w_{00}} +\frac{1}{w_{01}} +\frac{1}{w_{10}} +\frac{1}{w_{11}} })
\end{split}
\end{equation}


\subsection{MUTARA/HUNT}
\subsubsection{MUTARA/HUNT Background}
The algorithms Mining Unexpected Temporary Association Rules given the Antecedent (MUTARA) \cite{Jin2006} and Highlighting UTARs, Negating TARs (HUNT) \cite{Jin2010} were developed to be implemented on the Queensland Linked Data Set (QLDS) comprising of the Commonwealth Medicare Benefits Scheme (MBS), Pharmaceutical Benefits Scheme (PBS) and Queensland Hospital morbidity data. The QLDS contained hospital data from the 1 July 1995 to 30 June 1999 and MBS/PBS records from 1 January 1995 to 31 December 1999. Medical events from the hospital data were recorded by the International Statistical Classification of Diseases and Related Health Problems (ICD) 9 system and there were a total of 2020 different diagnoses. The drug prescriptions were coded with the World Health Organization (WHO) Anatomical Therapeutic Chemical (ATC) system \cite{ATC} and the database contained a total of 758 distinct drug codes. The QLDS did not have complete records for each patient and only contained hospital records.

\subsubsection{MUTARA/HUNT algorithms}
Both MUTARA and HUNT make use of a dependency measure known as the leverage that is frequently used when mining sequential patterns. In the context of this paper the leverage calculates the temporal dependency of medical event $C$ on drug $A$ by finding the number of patients that have drug $A$ followed by medical event $C$ within a time period of $T$ days (denoted $A \overset{T}{\rightarrow} C$) minus the number of patients you would expected to have the medical event after the drug if the drug $A$ and medical event $C$ occurred independently of each other. 

The algorithms apply a case control approach where patients prescribed the drug are referred to as users and patients that have never been prescribed the drug are referred to as non-users. The non-users are used to estimate the background rate that a medical event occurs. When implementing the algorithms the values $T_{e}, T_{c}, T_{r}, T_{b} \in \mathbbm{N}$ are input. 

\begin{figure*}[t]
\centering
\includegraphics[width=\textwidth]{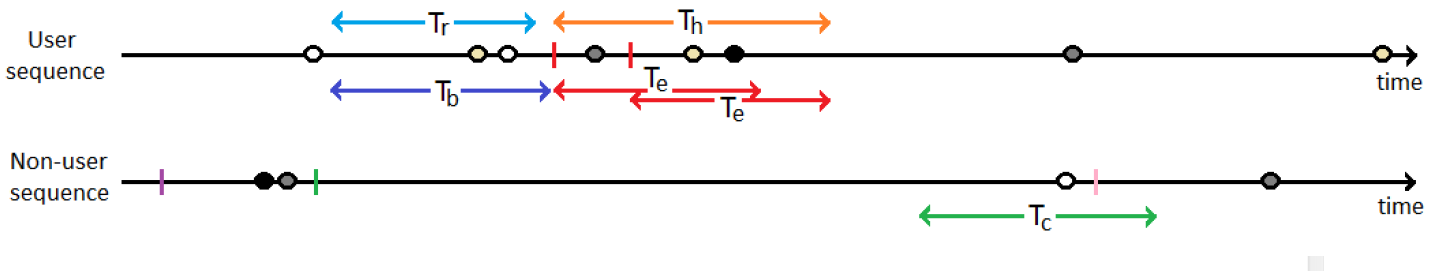}
\caption{Example of a user sequence and a non-user sequence with the time periods illustrated. The red vertical lines in the user sequence corresponds to the point in time that the user is prescribed the drug being investigated, non red vertical lines correspond to prescriptions of other drugs and the circles correspond to medical events, with different colours representing different medical events.}
\label{fig:MUTARA}
\end{figure*}

The algorithms first restrict their attention to subsequences of the user and non-user sequences. For each user sequence, the $T_{h}$ constrained subsequence of interest is the subsequence of length $T_{h}$ days starting from the day the drug is first prescribed. The value of $T_{h}$ differs between users depending on if the user has a repeat prescription within $T_{e}$ days after the first prescription. If the user does not have a repeat prescription within $T_{e}$ days of the first prescription then $T_{h}=T_{e}$, whereas if the second prescription of the drug occurs $s$ days after the first prescription where $s \le T_{e}$ then $T_{h}=s+T_{e}$. For each non-user, the $T_{c}$ constrained subsequence of interest is a subsequence of length $T_{c}$ days that is randomly chosen from the non-user's sequence. See Fig \ref{fig:MUTARA} for an illustration of how the subsequences are chosen.

We define $tot$ to be the number of users and non-users. Using the constrained subsequences, the $supp(A \overset{T}{\rightarrow} C )$ is defined as the number of user $T_{h}$ constrained subsequences containing the medical event $C$ divided by $tot$, the $supp(A \overset{T}{\rightarrow})$ is the number of users divided by $tot$ and $supp(\overset{T}{\rightarrow} C)$ is the number of user $T_{h}$ constrained subsequences that contain the medical event $C$ divided by $tot$ plus the number of non-user $T_{c}$ constrained subsequences that contain the medical event $C$ divided by $tot$. The leverage is calculated as,

\begin{equation}
Leverage = supp(A \overset{T}{\rightarrow} C)- supp(A \overset{T}{\rightarrow}) \times supp(\overset{T}{\rightarrow} C)
\label{eq:leverage}
\end{equation} 

\begin{figure*}[t]
\centering
\includegraphics[width=\textwidth]{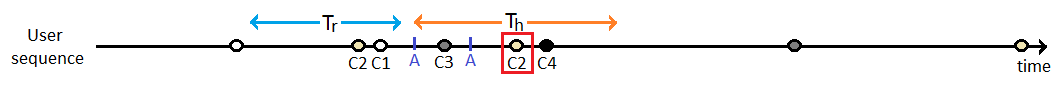}
\caption{Example of determining if an event is `predictable' or not within the sequence. If the event occurs in the hazard period $T_{h}$ and reference $T_{r}$ period it is `predictable', if it occurs in the hazard period and not the reference period it is not `predictable'. }
\label{fig:unexpfilt}
\end{figure*} 
In addition to calculating the standard leverage, a new measure called unexpected-leverage is also calculated. The unexpected-leverage (unexlev) makes use of a user's history to filter repeated medical events from the users's $T_{h}$ constrained subsequence as these are `predictable' and unlikely to be ADRs. This is done by investigating a reference period of length $T_{r}$ days starting from $T_{b}$ days prior to the first prescription within the user's sequence and filtering medical events from the user's $T_{h}$ subsequence if they occurred during the $T_{r}$ days in the period prior to the first prescription. An example of determining if an item is `predictable' can be seen in Fig. \ref{fig:unexpfilt} where three medical events $C2$,$C3$ and $C4$ all occur within the $T_{h}$ days after the prescription but the medical event $C2$ also occurred within the $T_{r}$ days prior to the prescription so it is considered to be `predictable' and is filtered, whereas the medical events $C3$ and $C4$ do not occur within the $T_{r}$ days prior so they are not filtered. Defining supp($A \overset{T}{\hookrightarrow} C$) as the number of users who's $T_{h}$ constrained subsequence contains medical event $C$ but who do not have medical event $C$ within the $T_{r}$ days prior to the first prescription divided by $tot$ and $supp(\overset{T}{\hookrightarrow} C)$ as the total of the number of users who's $T_{h}$ constrained subsequence contains medical event $C$ but who do not have medical event $C$ within the $T_{r}$ days prior to the first prescription plus the number of non-user $T_{c}$ constrained subsequences that contain the medical event $C$ all divided by $tot$, the unexpected leverage is calculated as,

\begin{equation}
unexlev = supp(A \overset{T}{\hookrightarrow} C)- supp(A \overset{T}{\rightarrow}). supp(\overset{T}{\hookrightarrow} C)
\label{eq:unleverage}
\end{equation} 

MUTARA was designed to calculate the unexpected-leverages for a drug and set of specified medical events. MUTARA then ranks the medical events in descending order of the unexpected-leverage, so the dependency measure used by MUTARA for signalling ADRs is the unexpected-leverage value. 

The authors state that MUTARA is prone to therapeutic failures so HUNT was developed as an improvement \cite{Jin2010}. HUNT returns medical events in descending order of the ratio between the leverage rank and the unexpected-leverage rank,
\begin{equation}
RankRatio = \frac{\mbox{event rank based on leverage}}{\mbox{event rank based on unexpected-leverage}}
\end{equation}
The medical events are ranked such that the medical event with the highest leverage/unexpected-leverage is ranked number $1$ and the higher the rank, the lower the rank number. The reason the rank ratio works is because the medical events that are true ADRs will generally only occur after the drug is taken and will not be filtered from the $T_{h}$ constrained subsequences, so the leverage and unexpected-leverage of ADRs will be approximately the same. The medical events that are linked to the cause of taking the drug will occasionally occur before the prescription for some patients and be filtered from the $T_{h}$ constrained subsequences, resulting in the unexpected-leverage being lesser than the leverage. Therefore, the rank number of ADRs based on unexpected-leverage will be less than or equal to the rank number based on leverage (as medical events linked to the cause of taking the drug will have moved down the rankings), so the rank ratio will be greater than or equal to $1$, whereas medical events linked to the cause of the drug will have an unexpected-leverage rank number greater than or equal to the leverage rank number, so the rank ratio will be less than or equal to $1$. It is clear that ADRs will have a higher $RankRatio$ than therapeutic failure medical events. 

\subsubsection{MUTARA/HUNT Previous Results}
MUTARA was applied on the QLDS for older females given alendronate and within the top ten events returned by the algorithm was reflux esophagitis, a known ADR. The top two events returned by MUTARA were linked to osteoporosis, the cause of taking the drug and the other events were not listed and assumed to be unlikely to be adverse drug events. This suggests the precision of the top ten events ($P(10)$) is $0.1$ for the algorithm applied to QLDS with a specific strata.

HUNT was applied on the QLDS for the drug alendronate and the stratum consisting of older females. The results show HUNT returned three known ADRs within the top ten events returned. This corresponds to an $P(10)=0.3$. When alendronate was investigated for older males, one known ADR was returned, corresponding to an $P(10)=0.1$.

\subsubsection{Implementation of MUTARA/HUNT in this study} 
In this study the set of medical events input in MUTARA and HUNT is the collection of any medical event that occurred within $30$ days of a first prescription of the drug being investigated for any patient. The reason behind choosing $30$ days is for consistency with the other algorithms in this study, enabling a fairer comparison. We did not implement any stratification and the chosen study period was all the records from $1900-2010$. The algorithms MUTARA$_{60}$ and HUNT$_{60}$ referred to in the continuation of this paper were MUTARA and HUNT implemented with $T_{c}=T_{e}=30$, $T_{r}=T_{b}=60$ and the algorithms MUTARA$_{180}$ and HUNT$_{180}$ referred to in the continuation of this paper were MUTARA and HUNT implemented with $T_{c}=T_{e}=30$, $T_{r}=T_{b}=180$. We chose not to have a gap between the reference period and date of first prescription of the drug ($T_{r}=T_{b}$) as preliminary tests indicated this gave better results due to many indicator events occurring a few days prior to prescription.


\subsection{TPD Ratio}
\subsubsection{TPD Ratio Background}
Noren {\it et al.} developed the Temporal Pattern Discovery (TPD) algorithm based on a measure of disproportionality \cite{Noren2010}. This method is very similar to the disproportionality methods applied to SRS databases. 

The database the TPD was developed for is the UK IMS Disease Analyzer containing over two million patients and 120 million prescriptions. Similar to the THIN database, the UK IMS Disease Analyzer extracts data directly from general practitioners computers. The OE ratio was implemented on patient records up to 31 December 2005. The database contained $3 445$ drugs and $5 753$ medical events encoded by the ICD-10 \cite{ICD10}. 

\subsubsection{ TPD algorithms}
The TPD algorithm compares the number of patients that have the first prescription of drug $x$ in thirteen months followed by event $y$ within a set time $t$ relative to the expected number of patients if drug $x$ and event $y$ were independent. Letting $n_{xy}^{t}$ denote the number of patients that have drug $x$ for the first time (in thirteen months) and event $y$ occurs within time period $t$, $n_{.y}^{t}$ denote the number of patients that are prescribed any drug for the first time (in thirteen months) and have event $y$ within time period $t$. $n_{x.}^{t}$ denote the number of patients that have drug $x$ for the first time (in thirteen months) with an active follow up in time period $t$ and $n_{..}^{t}$ denote the number of patients that have any drug for the first time (in thirteen months) with an active follow up in time period $t$. The expected number of patients that have drug $x$ and then event $y$ in a time period $t$ is then,
\begin{equation}
E_{xy}^{t} = n_{x.}^{t} \frac{n_{.y}^{t}}{n_{..}^{t}}
\label{eq:expec}
\end{equation} 
If for a given drug, the event occurs more than expected, the ratio between the observed and expected will be greater than one. By taking the $log_{2}$ of the ratio, a positive values suggests an interesting association between a drug and event. Modifying the equation to prevent the problem of rare events or drugs resulting in a small expectation that can cause volatility, a statistical shrinkage method is applied. 
\begin{equation}
IC = log_{2} \frac{n_{xy}^{t} + 1/2}{E_{xy}^{t} + 1/2}
\label{eq:icshrink}
\end{equation}
The shrinkage adds a bias for the $IC$ towards zero when an event or drug is rare. The credibility intervals for the $IC$ are the logarithm of the solution to Eq. \ref{eq:ci} with $q=0.025$ and $q=0.975$.
\begin{equation}
\int_{0}^{\mu_{q}} \frac{(E_{xy}^{t}+1/2)^{n_{xy}^{t}+1/2}}{\Gamma(n_{xy}^{t}+1/2)} u^{(n_{xy}^{t}+1/2)-1} e^{-(n_{xy}^{t}+1/2)} du = q
\label{eq:ci}
\end{equation}
The above can find possible drug and event associations of interest for a given $t$, however, the authors suggest that general temporal patterns can be found by comparing the $IC$ of two different time periods. The follow-up period of primary interest is denoted by $u$ and the control time period by $v$. This removes event and drug relationships that just happen to occur more in certain sub-populations. The difference between the $IC$ for both time periods is,
\begin{equation}
log_{2} \frac{n_{xy}^{u}}{E_{xy}^{u}} - log_{2} \frac{n_{xy}^{v}}{E_{xy}^{v}}
\end{equation} 
re-arranging and adding a shrinkage term gives,
\begin{equation}
IC_{\Delta} = log_{2} \frac{n_{xy}^{u}+1/2}{E_{xy}^{u*}+1/2}
\end{equation}
where
\begin{equation}
E_{xy}^{u*} = \frac{n_{xy}^{v}}{E_{xy}^{v}}.E_{xy}^{u} 
\end{equation}

In previous work, Noren {\it et al.} set $u$ and $v$ such that $30$ days after the prescription is contrasted with a time period of $27$ to $21$ months prior to prescription, this centers the $v$ period around the same time of the year that the $u$ period occurs preventing seasonal bias. The TPD calculates the $IC_{\Delta}$ for each medical event that occurs within a month of the drug being investigated for any patient and returns the list of the medical events in descending order of the $IC_{\Delta}$ value. In addition to calculating the $IC_{\Delta}$, a filter is applied to remove medical events with an $IC$ on the day of prescription or the month prior higher than the $IC$ for the time period investigated after the prescription.

\subsubsection{TPD Previous Results}
The algorithm was applied to the UK IMS Disease Analyzer database for the drug nifedipine and found seven known ADRs within the top ten non administrative events returned. This corresponds to an $P(10)=0.7$.

\subsubsection{Implementation of TPD in this study} 
In this study we implemented the TPD as described in \cite{Noren2010}, with $30$ days after the first prescription in $13$ months contrasted with $27$ to $21$ months prior to prescription, but investigated two different filters: 

\begin{itemize}
\item We apply the TPD and filter medical events with an $IC$ value the month prior to prescription or an $IC$ value on the prescription day greater than the $IC$ value during the month after the prescription (TPD 1).
\item We apply the TPD and filter medical events with an $IC$ value the month prior to prescription greater than the $IC$ value during the month after the prescription (TPD 2). 
\end{itemize}

The justification for choosing two filters is due to the possibility that ADRs can occur and be reported to doctors on the same day as the prescription, so filtering events with an $IC$ value on the day of prescription greater than the $IC$ value during the month after the prescription may prevent detection of some ADRs. We can investigate this by implementing both filters and comparing the results.


\section{Materials \& Methods} 
\subsection{THIN Database}
The THIN database contains medical records from participating general practices within the UK. The data are anonymously extracted directly from the general practice Vision clinical system \cite{inps2011}. THIN then implements validation steps; these are added as extra fields within the tables. The database contains patient information including the year of birth, gender, date of registration and family history of each patient registered at the practice since participation. Any illness, symptom, procedure, laboratory test, diagnosis or other relevant information that a doctor learns about a patient (referred to as a medical event) is recorded along with the corresponding date. Information regarding any medication prescribed as well as the date of the prescription and the dosage are also included in the database. For this comparison a database containing records from 495 general practices was used. This subset of the THIN database contained approximately four million patients, over 358 million prescription entries and over 233 million medical event entries.

Each medical event is recorded in the database by a reference code known as a Read code. The Read codes used in the THIN database are an independent system designed specifically for primary care but every ICD-9-CM (International Classification of Diseases, Ninth Edition, Clinical Modification) code (or analogues) have a corresponding Read code \cite{Shephard2011}. The UK general practice database has a hierarchal event Read code structure. The longer the Read code the greater the detail. For example `H32..00' may represent the event `Depression' whereas `H321.00' may represent `Depression due to medication'. The last two elements of the Read code represent events of the same type but described differently for example `G1...00' and `G1...01' may represent `had a chat with patient' and 'had a chin wag with patient' respectively. Only the first five elements of the Read codes were used in this study and Read codes corresponding to administrative events were ignored as they did not offer any information about ADRs. There were a total of 90 122 different five element Read codes used in this study. 

\subsection{Preprocessing the THIN Database}
Data miners applying algorithms to EHDs obtained from general practice records need to be careful with newly register patients. As patients can change general practices at any age, when they register they may have a history of events that a doctor needs to record. The term `registration event dropping' is used when historic or previously diagnosed events of newly registered patients are entered into the database. For example, when a new patient first visits their doctor they may inform the doctor of a previously diagnosed chronic illness such as `diabetes'. This medical event will then be input into the database with a date corresponding to the visit, rather than the actual date the patient was diagnosed with diabetes. As the dates recored for the `registration event drops' are frequently incorrect, including them in a research study will bias results. Research suggests that 'registration event dropping' is significantly reduced after a patient is registered for a year. To prevent 'registration event dropping' biasing the results in this study, the first 12 months of medical history after registration are ignored for each patient as justified in \cite{Lewis2005}.

As patients can move to a different practice at any time (or die), in this study we only include prescriptions into the study where the corresponding patient is still active for a minimum of 30 days after. This prevents the bias due to `under-reporting' of adverse drug events that may occur if a patient no longer attends the practice. The last date a patient is active is considered to be the maximum date of any record for the patient or the patient's date of death. 
\subsection{Drugs}

To investigate the robustness of each existing algorithm, we applied the existing algorithms to a range of drugs. To determine if certain algorithms perform optimally under specific conditions, we chose to study multiple drugs from the same drug families. The reason we chose multiple drugs from the same family is that the drugs would generally have similar indications and side effects, but may have different rates of prescription or be used by different groups of patients, for example the drug may be predominantly prescribed to the young or to females . Therefore, by comparing the algorithm performance between drugs of the same family we are controlling for patient medical state while investigating the effect that the attributes (average age, male proportion, total number of patients) of the group of patients prescribed each drug have on the algorithm. 

The drugs Ibuprofen, Ketoprofen, Fenoprofen and Celecoxib used in this study are all from the same drug family known as non-steroidal anti-inflammatory drugs (NSAIDs). These drugs are typically prescribed for continuous pain associated with inflammation and have a variety of common side effects including gastrointestinal disturbances, hypersensitivity reactions and depression. Rarer side effects include congestive heart failure, renal failure and hepatic failure. Elderly patients are more prone to side effects associated with NSAIDs. In this study the the drugs tended to be prescribed sightly more to females with the male proportion ranging from $0.335-0.405$ and to older patients, although Ibuprofen was prescribed to younger patients more than the other NSAID drugs. The NSAID drug prescribed the most was Ibuprofen with over a million first in 13 month prescriptions, whereas Fenoprofen was only prescribed 1225 times for the first time in 13 months, see Table \ref{drug:nsaids}. 

\begin{table*}[t]
\centering

\caption{Information about the NSAID drugs investigated in this paper. Total is the number of times the drug is prescribed for the first time in 13 months, age is the average age of the patients who are prescribed the drug for the first time in 13 months and male proportion is the number of patients that are male divided by the total number of patients who are prescribed the drug for the first time in 13 months.}
\label{drug:nsaids}
\begin{tabular}{ccccccc}
Drug & Total & TPD & MUTARA& ROR & Age & male proportion \\
\hline
celecoxib & 68036 & 62946 & 62100 & 63416 & 62.49 & 0.335 \\
ibuprofen & 1178163 & 1012555 & 858819 & 903415 &45.56 &0.405 \\
ketoprofen & 72946 & 65718 &61710& 63536 & 58.17 &0.375 \\
fenoprofen& 1255 & 1008 & 975& 1036 & 56.29 &0.404 \\
\end{tabular}
\end{table*}

The quinolones are a class of drugs used to treat bacterial infections such as respiratory track infections and urinary-track infections. Ciprofloxacin, levofloxacin, moxifloxacin, nalidixic acid and norfloxacin are drugs from the quinolone family that are investigated in this paper. The quinolones have many side effects, including tendon rupture. The average age of the patients prescribed the quinolones for the first time in 13 months was similar between all the drugs, around the late fifties. The male proportion shows that females are prescribed quinolones more than males, but this was more obvious for norfloxacin and nalidixic acid. Ciprofloxacin was the most prescribed quinolone and moxifloxacin was the least common, with only 1465 prescriptions. Table \ref{drug:quino} shows the information on the drugs from the THIN database.

\begin{table*}[t]
\centering

\caption{Information about the Quinolone drugs investigated in this paper. Total is the number of times the drug is prescribed for the first time in 13 months, age is the average age of the patients who are prescribed the drug for the first time in 13 months and male proportion is the number of patients that are male divided by the total number of patients who are prescribed the drug for the first time in 13 months.}
\label{drug:quino}
\begin{tabular}{ccccccc}
Drug & Total & TPD & MUTARA& ROR & Age & male proportion \\
\hline
ciprofloxacin & 280011 & 250158 & 227739 &235420 & 55.64 &0.440 \\
levofloxacin & 7662 & 7028 &6775& 6928 & 60.55 & 0.43\\
norfloxacin& 14876 & 13224& 12220& 12625 & 56.83& 0.262 \\
moxifloxacin &1465 & 1347 &1343 &1371 & 62.09& 0.419 \\
nalidixic acid& 4273 & 3646& 3620& 3787& 55.63 & 0.127 \\
\end{tabular}
\end{table*}

Tricyclic antidepressant drugs are a family of drugs used to treat depression and are known to cause, among others, cardiovascular and central nervous system side effects. The three drugs, doxepin, lofepramine and nortriptyline where selected in this paper. The tricyclic antidepressants investigated are prescribed to patients with similar ages and genders and tend to be prescribed more often to older females. The main difference between the drugs is that doxepin is only prescribed to $6752$ patients whereas the other two drugs are prescribed to more than $10 000$ patients, see Table \ref{drug:tri}. 
\begin{table*}[t]
\centering

\caption{Information about the tricyclic drugs investigated in this paper. Total is the number of times the drug is prescribed for the first time in 13 months, age is the average age of the patients who are prescribed the drug for the first time in 13 months and male proportion is the number of patients that are male divided by the total number of patients who are prescribed the drug for the first time in 13 months.}
\label{drug:tri}
\begin{tabular}{ccccccc}
Drug & Total & TPD & MUTARA& ROR & Age & male proportion \\
\hline
doxepin &6752 & 6029& 5908& 6104& 56.69 &0.316\\
lofepramine &45532 & 38565& 37642& 39517& 51.39& 0.285 \\
nortriptyline &11775 & 10519& 10307& 10650& 54.43& 0.286 \\
\end{tabular}
\end{table*}

The drugs nifedipine, nicardipine, amlodipine, felodipine and verapamil are all calcium channel blocker that are used to treat high blood pressure and raynaud's phenomenon. It is common for the calcium channel blockers to be prescribed with other drugs and applying the existing algorithms to detect side effects on the calcium channel blockers will investigate the effect of confounding due to multiple prescriptions. The drug nifidipine was previously used to investigate the TPD applied to the UK IMA Disease Analyzer, so investigating the calcium channel blockers will also give insight into how robust the TPD is when applied to different electronic healthcare databases. The calcium channel blockers are generally prescribed for the first time in 13 months to patients around 65 years old. Amlodipine and nicardipine are prescribed only slightly more to females than males, whereas the other calcium channel blockers investigated are prescribed even more often to females. Amlodipine and nifedipine have been prescribed over $100 000$ times for the first time in 13 months in the THIN database, but nicardipine has only been prescribed $2796$ times, see Table \ref{drug:ccb}.

\begin{table*}[t]
\centering
\caption{Information about the calcium channel blocker drugs investigated in this paper. Total is the number of times the drug is prescribed for the first time in 13 months, age is the average age of the patients who are prescribed the drug for the first time in 13 months and male proportion is the number of patients that are male divided by the total number of patients who are prescribed the drug for the first time in 13 months.}
\label{drug:ccb}
\begin{tabular}{ccccccc}
Drug & Total & TPD & MUTARA& ROR & Age & male proportion \\
\hline
nifedipine &125491 &112715 & 112499 &115823 & 65.29& 0.453\\
verapamil &24334 & 22000& 21896 &22513 & 65.01& 0.405\\
felodipine &69534 & 65093 &64036 &65202 & 67.46& 0.454\\
amlodipine& 270918 & 251316 &249972 &254876 & 66.68& 0.494\\
nicardipine& 2796 & 2510 &2511& 2593 & 65.91& 0.481\\
\end{tabular}
\end{table*}

The sulphonylurea drug family includes tolbutamide, glibenclamide, gliclazide, glimepiride and glipizide. They are a class of antidiabetic drugs used for the management of type 2 diabetes mellitus. The sulphonylureas are prescribed for the first time in 13 months to older patients will an average age around 65 years old and all the sulphoylureas investigated except tolbutamide are prescribed more often to males, with approximately equal male proportions. Glipizide and tolbutamide are the less frequently prescribed sulphonylurea drugs. The general information about each of the sulphonylurea drugs can be seen in Table \ref{drug:sul}.

\begin{table*}[t]
\centering
\caption{Information about the sulphonylurea drugs investigated in this paper. Total is the number of times the drug is prescribed for the first time in 13 months, age is the average age of the patients who are prescribed the drug for the first time in 13 months and male proportion is the number of patients that are male divided by the total number of patients who are prescribed the drug for the first time in 13 months.}
\label{drug:sul}
\begin{tabular}{ccccccc}
Drug & Total & TPD & MUTARA& ROR & Age & male proportion \\
\hline
glibenclamide &11874 & 10356& 10377& 10768& 65.12& 0.540\\
gliclazide &45824 & 41626& 40537& 41612& 65.02& 0.546\\
glimepiride& 10957 & 10156& 9882& 10081& 64.20& 0.534\\
glipizide& 5315 & 4856& 4614& 4731& 66.50& 0.535\\
tolbutamide &3113 & 2758& 2793& 2894& 69.40&0.487\\
\end{tabular}
\end{table*}

The last drug family is the Penicillin drugs amoxicillin, ampicillin, flucloxacillin, benzylpenicillin and phenoxymethlypenicillin. These drugs are used to treat bacterial infections. The number of times the drugs are recorded as being prescribed in the THIN database varies between $2000$ to over two million. There is also a divergence between the average age of the patients prescribed each of the drugs, with the penicillins generally being prescribed to younger patients than many of the other drugs families investigated in this paper. The male proportion is fairly similar between the different penicillin drugs, with females being prescribed the drug more often than males, see Table \ref{drug:pen}. 
\begin{table*}[t]
\centering
\caption{Information about the penicillin drugs investigated in this paper. Total is the number of times the drug is prescribed for the first time in 13 months, age is the average age of the patients who are prescribed the drug for the first time in 13 months and male proportion is the number of patients that are male divided by the total number of patients who are prescribed the drug for the first time in 13 months.}
\label{drug:pen}
\begin{tabular}{ccccccc}
Drug & Total & TPD & MUTARA& ROR & Age & male proportion \\
\hline
amoxicillin& 2795759 & 2321098& 1593874 &1718875& 38.84& 0.427\\
benzylpenicillin& 2071 & 1610& 1840& 1972& 31.79& 0.471\\
flucloxacillin& 971174 & 834017& 729967& 765428& 41.42& 0.456\\
phenoxymethly& 55397 & 45941& 45679& 48142& 29.67& 0.396\\
ampicillin &80655 & 63458& 64827& 69381& 39.18& 0.423\\
\end{tabular}
\end{table*}

\subsection{Comparison Method}
Before calculating the comparison measures we applied our own filter that is specific to the THIN database to remove medical events that are chronic (as chronic medical events are not side effects), that correspond to cancer (as it is unlikely that cancer will develop as a side effect within 30 days of the drug first being prescribed) or those that are not related to the patient's health (such as administration events or occupation information). Without this filter the performance of each algorithm was reduced. 

\subsubsection{Natural threshold based measures}
The signalling ability of each algorithm using the natural threshold was investigated by calculating the sensitivity and specificity at their natural thresholds (0 for MUTARA/TPD and 1 for ROR) using the values described in Table \ref{tab:tp}.
\begin{table}
\centering
\caption{ Table illustrating a TP, FP, FN and TN.  Listed event means the medical event is listed on the British National Formulary as a known side effect and non-listed event means it is not listed as a known side effect.  }
\label{tab:tp}
\begin{tabular}{c|cc}
& Listed event & Non-listed event \\ \hline
Signalled & True Positive (TP) & False Positive (FP) \\
Not Signalled & False Negative (FN) & True Negative (TN) \\
\end{tabular}
\end{table}

\begin{equation}
\mbox{Sensitivity} = \frac{TP}{TP+FN}
\end{equation}

\begin{equation}
\mbox{Specificity} = \frac{TN}{TN+FP}
\end{equation}

\subsubsection{Rank based measures}
Each data mining algorithm, as described in section \ref{sec:1}, is applied to the THIN database and a ranked list of medical events is returned for each drug investigated. The non filtered medical events are ranked in descending order of the association between the drug of interest and medical event, so medical events the algorithm has deemed more likely to be ADR are ranked higher, an example of this can be seen in Table \ref{tab:list}. Each algorithm is then analysed by investigating how well it has ranked each of the known ADRs that have occurred in the returned ranked list. The known ADRs are those that are listed in the British National Formulary (BNF) \cite{BNF} for the specific drug, or medical events described as `adverse reaction to drug x' or containing information about the continuation of the drug prescription. 

Given a ranked set of events, we calculate the Truth measure $y_{(i)}$ for the $i^{th}$ ranked event, by letting $y_{(i)}= 1$ if the event is a known ADR and $y_{(i)}=0$ otherwise, as shown in Table \ref{tab:list}. Table \ref{tab:list} shows an example of a returned list containing five events ranked by an algorithm and the corresponding $y$ values. Using the $y$ values we can then use the measures described below to compare the different algorithms. 

\begin{table*}[t]
\centering
\caption{An example of the medical event list associated to a specific drug and ordered by one of the algorithms.}
\label{tab:list}

\begin{tabular}{ccccc}
Medical Event & Rank Score & Known ADR & $y_{(i)}$ & \\ \hline
Event 1 & 2.34 & No & $y_{(1)}= 0$ & \\
Event 5 & 2.12 & Yes & $y_{(2)}= 1$ & precision$_{2}=1/2$ \\
Event 4 & 1.75 & Yes & $y_{(3)}= 1$ & precision$_{3}=2/3$ \\
Event 2 & 1.74 & No & $y_{(4)}= 0$ & \\
Event 3 & 0.68 & No & $y_{(5)}= 0$ & \\
\end{tabular}
\end{table*}

The precision of each method at cutoff $K$, denoted $P(K)$, is defined as the fraction of known ADRs that occur in the top $K$ events of the list returned by each algorithm for a specific drug, see Eq. (\ref{acc}).
\begin{equation}
P(K)= \frac{\sum_{i=1}^{K}y_{(i)} }{K}
\label{acc}
\end{equation}

The average precision (AP) is a measure that can be used to determine how well an algorithm generally ranks the medical events associated to a drug. This measure has previously been applied to compare algorithms implemented on a different EHD \cite{Ryan2012}. 

The AP is calculated by finding the average $P(K)$ for each $K$ corresponding to a known ADR, 
\begin{equation}
\mbox{AP}= \frac{\sum_{K:y_{(K)}=1} P(K)}{\sum_{i} y_{(i)}}
\label{map}
\end{equation}
Using Table \ref{tab:list} as an example, as there are two known ADRs returned ($\sum_{i} y_{(i)}=2$) and the known ADRs in the table are ranked second and third we have $\{K:y_{(K)}=1\}=\{2,3\}$, so the AP score is, 

\begin{equation}
\mbox{AP}=\frac{P(2)+P(3)}{2}= \frac{1/2+2/3}{2}= \frac{7}{12}
\end{equation} 

It was also possible to investigate how well each algorithm ranks the known adverse drug events for a specific drug depending on how common they are. As the BNF states the risk of each known ADR by separately listing frequently, less frequently and rarely occurring known ADRs we also calculated the AP score of each algorithm when only considering rarely occurring known ADRs (as unknown ADRs are likely to be rare). So, we calculate the AP scores for two different situations; considering all known ADRs and considering only rarely occurring known ADRs. When considering only the rare ADRs, the known common and less common ADRs are filtered from the list of medical events returned by each algorithm. The reason for filtering the common and less common known ADRs is to prevent their presence causing a low AP score when only considering rare ADRs, as if an algorithm did not filter the common and less common known ADRs and correctly ranks these highly (above the rare ADRs) then the rare ADRs would have a lower rank, which would result in a lower AP score when only considering rare ADRs.

\begin{table*}
\centering
\caption{An example of the medical event list for all the drugs and ordered by one of the algorithms.}
\label{tab:comb}
\begin{tabular}{ccccc}
Drug & Medical Event & Rank Score & Known ADR & $y_{(i)}$ \\ \hline
Drug 10 & Event 7 & 2.34 & No & $y_{(1)}= 0$ \\
Drug 10 & Event 5 & 2.12 & Yes & $y_{(2)}= 1$ \\
Drug 2 & Event 56 & 1.75 & Yes & $y_{(3)}= 1$ \\
Drug 9 & Event 7 & 1.74 & No & $y_{(4)}= 0$ \\
Drug 2 & Event 16 & 0.68 & No & $y_{(5)}= 0$ \\
\end{tabular}
\end{table*}

To give a general measure of the ranking ability of each algorithm over all the drugs investigated (rather than calculating ranking measures per drug) we also compute ROC plots. The ROC plots were generated by combining all the drug results for each algorithm, as illustrated in Table \ref{tab:comb} and calculating the sensitivity and specificity for signals generated at a range of signalling thresholds. The ROC curves are formed by plotting the true positive rate (TPR=sensitivity) against the false positive rate (FPR=(1-specificity)). The Area Under the Curve (AUC) was approximated using the trapezoidal rule for FPR ranging between 0-1, 0-0.3 and 0-0.1 ($AUC_{[0,1]}$,$AUC_{[0,0.3]}$ and $AUC_{[0,0.1]}$ respectively). 

\section{Results}
\subsection{Natural Thresholds}
Table \ref{res:ss} shows the specificity and sensitivity for the different algorithms at their natural threshold and the number of signals generated. The natural threshold for the ROR and TPD are that their lower confidence interval value is greater than 1 and 0 respectively. MUTARA does not have a lower confidence interval calculation, so the natural threshold implemented is that the unexpected-leverage is greater than 0 and for HUNT the top 10\% of medical events were signalled. 

\begin{table}
\centering
\caption{The specificity and sensitivity at the natural thresholds for the different algorithms (3dp).}
\label{res:ss}
\begin{tabular}{ccccc}
Algorithm & Signals & Sens & Spec & Precision \\ \hline
HUNT$_{60}$ & 7785 &0.179 &0.903& 0.0541 \\
HUNT$_{180}$ & 7785& 0.193& 0.903& 0.058 \\
MUTARA$_{60}$ & 67624 & 0.933& 0.109& 0.032 \\
MUTARA$_{180}$ & 65435& 0.914& 0.136 &0.032 \\
TPD 1 & 1893& 0.090& 0.953& 0.057\\
TPD 2 & 3557& 0.107& 0.926& 0.043\\
ROR$_{05}$ & 37729 & 0.312 & 0.726 & 0.031 \\
\end{tabular}
\end{table}

\begin{table}
\centering
\caption{The AUC results for the different algorithms (3dp).}
\label{res:auc}
\begin{tabular}{cccc}
Algorithm & $AUC_{[0,1]}$ & $AUC_{[0,0.3]}$ & $AUC_{[0,0.1]}$ \\ \hline
HUNT$_{60}$ & 0.566 & 0.072 & 0.011 \\
HUNT$_{180}$ &0.570 &0.071 &0.011 \\
MUTARA$_{60}$ &0.596 &0.076 &0.010 \\
MUTARA$_{180}$ &0.597 &0.069 &0.010 \\
TPD 1 &0.570 &0.065 &0.009 \\
TPD 2 &0.557 &0.060 &0.007 \\
ROR$_{05}$ &0.546 &0.048 &0.005
\end{tabular}
\end{table}

\subsection{ROC Analysis} 
\begin{figure*}
\centering
\includegraphics[trim=0 15 30 50, clip,width=0.9\textwidth]{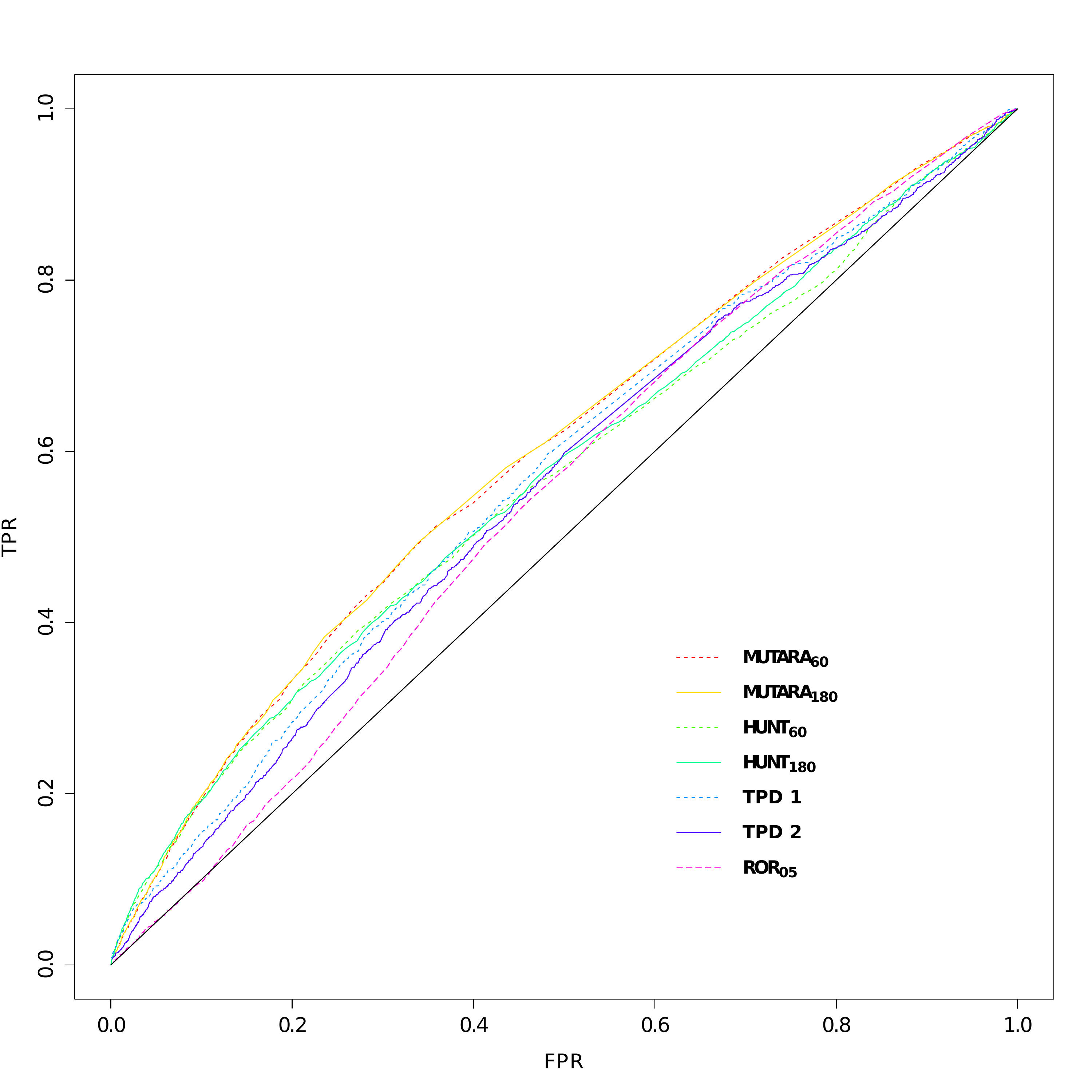}
\caption{The ROC plots for the different algorithms over the whole FPR (1-specificity) range. The black line is the line x=y.}
\label{fig:ror_all}
\end{figure*}

\begin{figure*}
\centering
\includegraphics[trim=0 15 30 50, clip,width=0.9\textwidth]{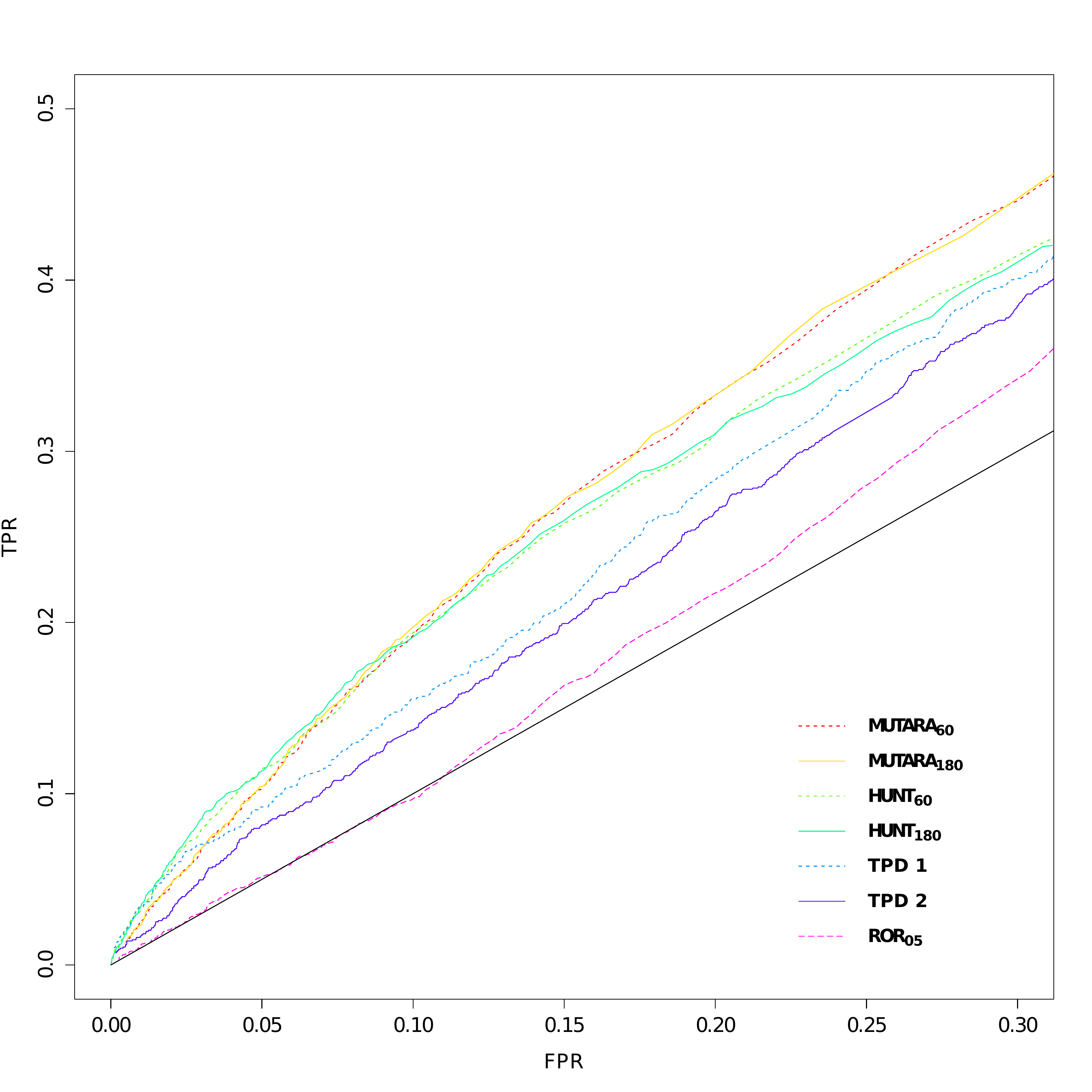}
\caption{The ROC plots for the different algorithms over the section of FPR (1-specificity) less than 0.3. The black line is the line x=y.}
\label{fig:ror_03fpr}
\end{figure*}

THE AUC$_{[0,1]}$ for the algorithms ranged between 0.546 (ROR$_{05}$) to 0.597 (MUTARA$_{180}$), the AUC$_{[0,0.3]}$ for the algorithms ranged between 0.048 (ROR$_{05}$) to 0.076 (MUTARA$_{60}$) and AUC$_{[0,0.1]}$ for the algorithms ranged between 0.005 (ROR$_{05}$) to 0.011 (HUNT$_{180}$ and HUNT$_{60}$). Fig. \ref{fig:ror_all} and Fig. \ref{fig:ror_03fpr} show the ROC plots for the different algorithms.

\subsection{AP}

\begin{figure*}
\includegraphics[width=\textwidth]{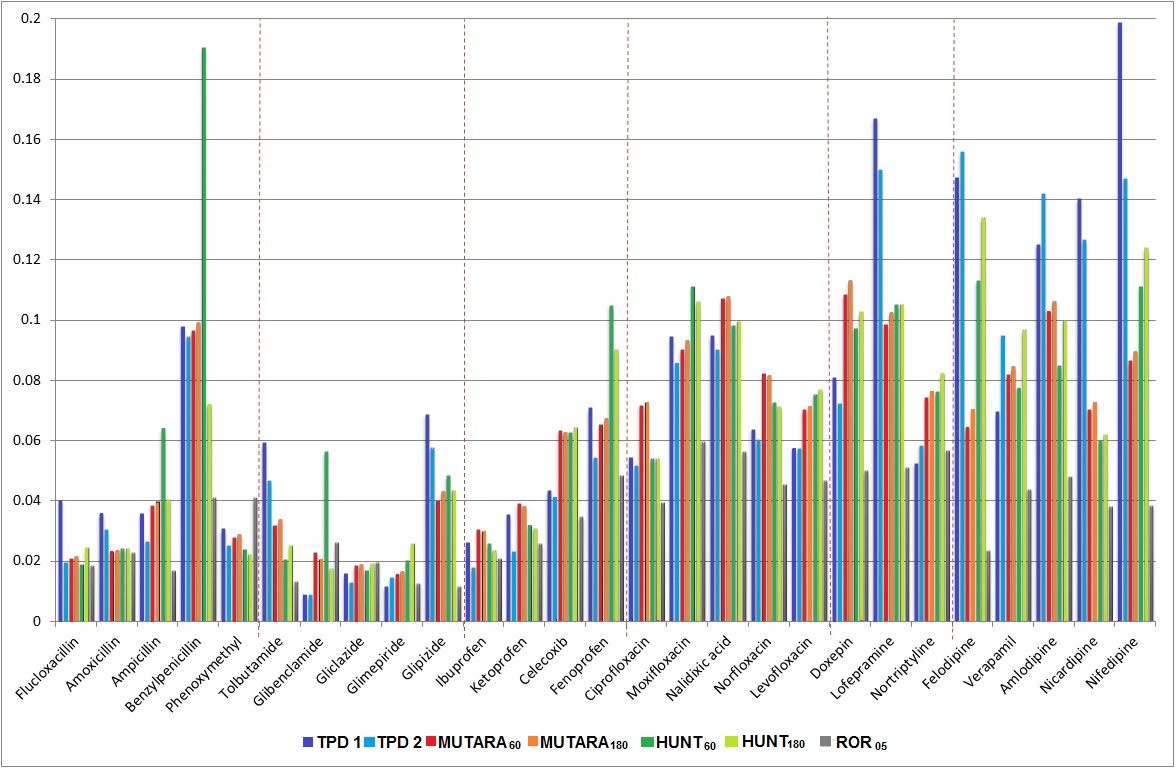}
\caption{Bar chart showing the AP scores for all the drugs, with a dashed line separating the different drug families.}
\label{fig:mal_all}
\end{figure*}

Fig. \ref{fig:mal_all} shows the AP scores for the different algorithms over the range of drugs investigated. The family of drugs that the algorithms perform worse on overall were the sulphonylureas with AP scores ranging from $0.0088-0.0687$. The algorithms all performed well on the calcium channel blockers, with AP scores ranging from $0.0236-0.1988$, but the ROR$_{05}$ performed worse for all the calcium channel blockers investigated. The algorithms also performed well for the tricyclic antidepressants with AP scores ranging between $0.0499-0.1670$. It can be seen in Fig. \ref{fig:mal_all} that generally the algorithms perform similarly between the same drugs of the same class, apart from the algorithms performing much better for benzylpenicillin sodium compared to the other penicillin drugs. 

\begin{figure*}
\includegraphics[width=\textwidth]{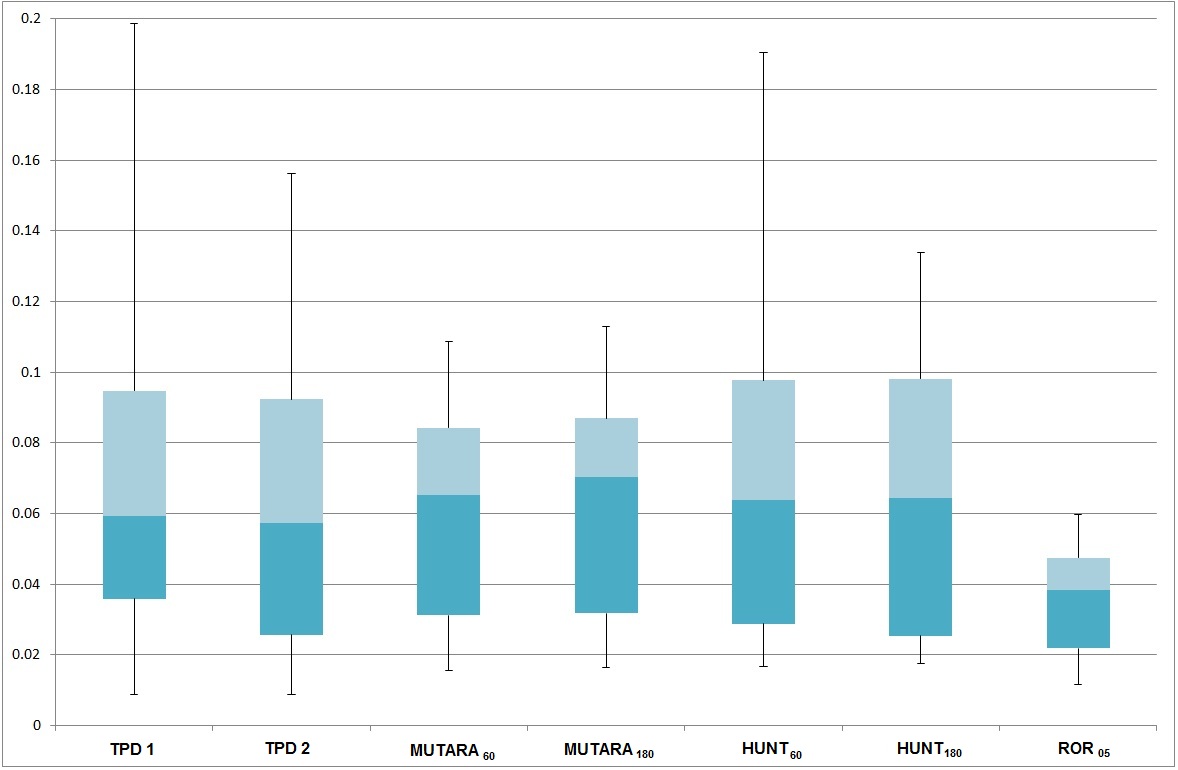}
\caption{Box plot showing the median, quartiles and minimum/maximum AP scores for each algorithm applied to all the drugs.}
\label{fig:box_all}
\end{figure*}

\begin{figure*}
\includegraphics[width=\textwidth]{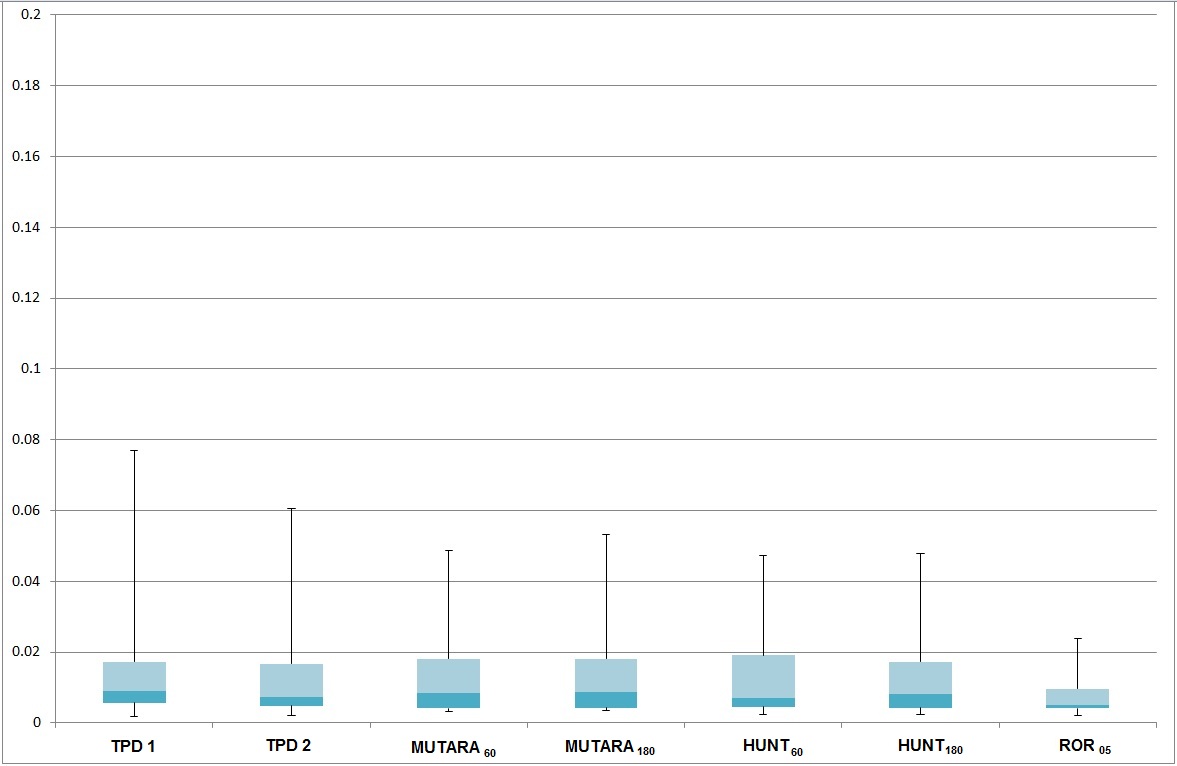}
\caption{Box plot showing the median, quartiles and minimum/maximum AP scores for each algorithm when only considering rare known ADRs for all the drugs.}
\label{fig:box_rare}
\end{figure*}

The box plots of the AP scores for the different algorithms seen in Fig. \ref{fig:box_all} show overall the TPDs, MUTARAs and HUNTs perform equally and outperform the ROR$_{05}$. The MUTARA algorithm has the highest median AP score over all the drugs and is more consistent, whereas the performance of the TPD and HUNT varies more between the drugs. The box plots for the algorithm AP scores when only considering rare known ADRs, see Fig. \ref{fig:box_rare}, shows that the algorithms AP scores are lower when ranking rare ADRs, as the maximum AP score for all the algorithms when only considering rare known ADRs was less than $0.08$, whereas the maximum AP scores for all the algorithms when considering all known ADRs was approximately $0.2$.

\section{Discussion}
\subsection{Natural Thresholds}
The results show that the natural thresholds operate at different stringencies. The most stringent algorithm was the TPD 1 that returned 1893 signals, the lowest out of all the algorithms, with a high specificity of 0.953 and low sensitivity of 0.09, whereas the less stringent was the MUTARA$_{60}$ that returned 67624 signals with a high sensitivity of 0.933 and a low specificity of 0.109. This was not unexpected as the TPD threshold used the lower confidence interval value rather than the actual $IC_{\Delta}$ value and the TPD applied a statistical shrinkage. The results also show that none of the algorithms was able to signal the known ADRs without being swamped by false positives. 
\subsection{ROC Analysis}
The existing EHD algorithms can be implemented to return a list of medical events in order of how likely the algorithm has deemed each medical event to be a potential ADR. In effect, the algorithms act as filters where lowly ranked medical events are filtered out and tentative signals are generated for the top $n$ ranked medical events. These tentative signals are investigated further with more stringent statistical analysis required to confirm if they are true signals or not. Therefore, to discover ADRs, the algorithms need to rank medical events corresponding to ADRs highly and the higher they are ranked the more likely they are to be discovered. This shows that AUC$_{[0,0.1]}$ and AUC$_{[0,0.3]}$ give a direct indication into how likely the algorithms will detect ADRs. As the natural threshold's of the algorithms act at different stringencies, the AUC$_{[0,0.1]}$ and AUC$_{[0,0.3]}$ measures offer a fairer comparison between the algorithms. 

The AUC results show that the algorithm perform similarly and no algorithm had a higher AUC for all three FPR cutoff values studied (AUC$_{[0,1]}$,AUC$_{[0,0.3]}$ and AUC$_{[0,0.1]}$).

\subsection{AP}
Overall no algorithm consistently outperformed the others over all the drugs investigated in this study, however, either the TPD 1 or HUNT had the highest AP score for the majority of the drugs studied. The ROR$_{05}$ generally performed the worse, but still had a higher AP score than the other algorithms for the drug phenoxymethylpenicillin. As we know the number of times each drug was prescribed, the average age of the patients prescribed each drug and the male proportion of the patients prescribed the drug, we discuss how these factors impacted on the different algorithms below.

\subsubsection{Age}
The drugs prescribed to the youngest patients were the penicillins. The TPD 1 returned the highest AP score for flucloxacillin, amoxicillin and phenoxymethylpenicillin and HUNT$_{60}$ returned the highest AP score for benzylpenicillin sodium and ampicillin. This is consistent with the general results as the TPD 1 and HUNT$_{60}$ generally had the highest AP score over all the drugs. Although, compared to the other drugs in the study all the algorithms performed fairly poorly for the penicillins suggesting drugs that on average are prescribed to younger patients may require the algorithms to apply stratification techniques rather than applying the algorithms to all ages. This is unfortunate as stratification is likely to result in an increase in the time it takes to identify an ADR, as the population of patients being prescribed a drug needs to be partitioned by age but there is a lower limit to the patient size required for the algorithms to work accurately. 

The drugs prescribed to the oldest patients were tolbutamide, felodipine, amlodipine, glipizide and nicardipine with the average age of the patients prescribed the drugs being 69, 67, 67, 67 and 66 respectively. These drugs had high AP scores and the TPD performed the best for all of these drugs. This suggests the TPD is the optimal algorithm to apply to the THIN database if the drug is prescribed to older patients. One possible reason why the TPD is better at ranking ADRs for older patients may be due to older patients experiencing more medical events, so there are more medical events to rank, but as the TPD applies a filter it can remove many of these medical events, increasing the chance of ranking a known ADR highly, whereas the other algorithms apply less stringent filters.

\subsubsection{Gender Bias}
The drugs with the greatest gender bias towards females (prescribed more often to females) were the drugs norfloxacin, lofepramine , nortriptyline and doxepin with male proportions of 0.26,0.28, 0.29 and 0.32 respectively. Apart from lofepramine, MUTARA and HUNT perform much better for the drugs prescribed more to females than males. This may highlight the need for MUTARA and HUNT to be applied to a subset of the patients such as old females or young males as was implemented originally for MUTATA and HUNT.

The drugs gliclazide, glibenclamide, glipizide and glimepiride are prescribed more often to males than females with a male proportion of 0.55, 0.54, 0.53 and 0.53 respectively. These drugs had the lowest AP scores compared to all the other drugs investigated and possibly suggests that the algorithms have difficulties detecting ADRs for drugs prescribed more often to males. This result is not unexpected as patients taking part in clinical trails are predominately male, so the known ADRs for drugs frequently prescribed to males are more likely to be rare (otherwise they would be discovered during the clinical trials).

\subsubsection{Low Prescription Rates}
The drugs that had the lowest number of prescriptions in the database were fenoprofen, moxifloxacin, benzylpenicillin and tolbutamide with only $1255$, $1465$, $2027$ and $3113$ prescriptions respectively. These four drugs had a varied average age of patients prescribed the drugs and male proportion. Looking at Fig \ref{fig:mal_all} we can see that the AP scores for these drugs are the best or second best when comparing against the drugs in the same family. This result shows that the existing algorithms applied to the THIN database are able to detect ADRs after around one thousand patients have been prescribed a drug and may be able to detect ADRs in new drugs more efficiently that applying existing algorithms to SRS databases. Furthermore, HUNT performed the best for the three drugs with the lowest number of prescriptions, suggesting HUNT may be the optimal algorithm to apply when a drug is newly marketed.

\subsection{Detecting rare ADRs}
The results show that none of the existing algorithms is able to rank rare ADRs highly and currently the existing algorithms tend to focus on detecting more common ADRs. A possible reason for this with the TPD is because it is more stringent as the calculation is biased to reduce the $IC_{\Delta}$ when a medical event is rare, this means that we can be more confident about signals generated by the TPD, but at the cost that rarer ADRs may take longer to be signalled. The reason MUTARA and HUNT struggle to rank rare ADRs highly is likely due to non ADR medical events that are linked to the cause of the drug incorrectly having a high rank and pushing the rare ADRs down the ranked list.

This is an important result as the currently unknown ADRs are either likely to be rare or correspond to medical events with a high background rate, as more obvious ADRs will be discovered during clinical trails or by mining SRS databases.

\subsection{ Discussion of Existing Algorithms}
There does not appear to be an optimal reference period to apply for HUNT out of $T_{r}=60$ or $T_{r}=180$ as HUNT$_{180}$ returned a better AP score that HUNT$_{60}$ for some drug and worse for others. This shows the reference period used by HUNT has a large impact on the outcome and should be different for each drug. The reason for this is that the reference period should depend on the background rate of the drugs ADRs, as if the ADR is common, having a large reference period will mean that the ADRs is likely to occur during the reference period and be incorrectly filtered. However, these is also a downside to having a short reference period as the shorter the reference period the less likely the algorithm will filter medical events have are repeats. An improvement to MUTARA and HUNT would be to develop a way to learn the optimal reference period to use for each drug. MUTARA and HUNT also have issues due to using a random period of time in the non-user sequences making the algorithms non deterministic and in this paper we did not study how much the results vary between different implementations of the algorithms.

We investigated the two different filters for the TPD and interestingly including the day of prescription into the filter improved the AP scores for the majority of drugs, this suggests that most ADRs are not recorded on the same day as the drug is prescribed. However, for some drugs the TPD 2 outperformed the TPD 1, so using the day of prescription in the filter does occasionally prevent the detection of known ADRs. In summary, including the day of prescription into the filter for the TPD generally increases the algorithms precision but at a cost as it can prevent the detection of some ADRs. 

\subsection{Previous Work and Limitations}
The results obtained in this study were consistent with previous results as the $P(10)$ for MUTARA and HUNT averaged $0.065$ and $0.122$ respectively in this study and were $0.1$ and $0.1-0.3$ respectively in previous work \cite{Jin2006}\cite{Jin2010}. The $P(10)$ for the TPD method applied to Nifedipine in this study was $0.7$, the same as on the UK IMS Disease Analyzer database \cite{Noren2010}. However, there was deviation between the AP score of the $ROR_{05}$ in this study ($0.01-0.06$) and in the study by Zorych {\it et al.} \cite{Zorych2011} (0.1-0.15), this is probably due to this study using real data with redundant Read codes and Zorych {\it et al.} using simulated data. 

One limitation with this study is that there is no `gold standard' for ADR detection and it is impossible to calculate the true sensitivity, specificity and AP scores of the algorithms as the complete set of ADRs for each drug is unknown and the algorithms may correctly rank an unknown ADR highly but this would lower the sensitivity, specificity and AP values in this study due to the event not being listed by the BNF as a known ADR. However, the algorithms should be able to correctly rank the known ADRs and these are likely to be more common and obvious, so if the algorithm is unable to correctly rank these above noise events then it is unlikely to identify the unknown ADRs, so the AP scores determined in this study still give insight into the algorithms abilities to detect ADRs. The redundancy in the Read Codes may also result in biased results as similar medical events with different Read Codes may get ranked closely and push down the ranks of known ADR, so the algorithms may have higher AP scores if there was a way to group Read Codes corresponding to the same medical event. 

\section{Conclusions}
\label{sec:conclusion}
In this paper we compared four existing ADR detecting algorithms by applying them to the THIN database for a range of drugs and measured how well they ranked the known ADRs or signal known ADRs at their natural thresholds. We have determine that the benchmark AP score for ADRs signalling algorithms applied to the THIN database is 0.2, the benchmark AUC$_{[0,0.1]}$ is 0.011 (3dp) and AUC$_{[0,0.3]}$ is 0.076 (3dp). Future algorithms should aim for higher scores. The results show that no algorithm was superior for all the drugs considered, although the results do indicate that HUNT may be the optimal algorithm to apply when the number of patients prescribed the drug is low or the TPD may be optimal when the patients are old, however further research needs to be conducted to confirm these hypotheses. However, in many situations at present, the tentative ADR signals should be generated based on how the medical events are ranked by all the different algorithms rather than relying on one algorithm alone.

The results also clearly show that the existing algorithms are not capable of detecting rare ADRs, but one of the potential advantages of mining the THIN database compared with mining the SRS databases it the ability to detect rare ADRs that are under-reported in SRS databases. For example, some rare but fatal ADRs might never be noticed and reported into an SRS database and would therefore be undetectable by mining SRS databases.

Future work could focus on developing new filters that are able to remove the medical events that are related to the cause of taking the drug but do not occur before the drug is prescribed, such as illness progression events.  This would then reduce the number of false positive associations that are returned by the algorithms. Future work should also address the limitations of this study by developing a way to cluster the Read codes and reduce their redundancy.

%


\bibliographystyle{spmpsci} 
\bibliography{LitRevRef2_3authors,LitRevRef_extras} 

\begin{thebibliography}{10}
\providecommand{\url}[1]{{#1}}
\providecommand{\urlprefix}{URL }
\expandafter\ifx\csname urlstyle\endcsname\relax
  \providecommand{\doi}[1]{DOI~\discretionary{}{}{}#1}\else
  \providecommand{\doi}{DOI~\discretionary{}{}{}\begingroup
  \urlstyle{rm}\Url}\fi

\bibitem{Almenoff2005}
Almenoff, J., Tonning, J.M., Gould, A.L., et~al: Perspectives on the use of
  data mining in pharmacovigilance.
\newblock {\it Drug Saf} \textbf{28(11)}, 981--1007 (2005)

\bibitem{Alvarez1998}
Alvarez-Requejo, A., Carvajal, A., Begaud, B., et~al: Under-reporting of
  adverse drug reactions- estimate based on a spontaneous reporting scheme and
  a sentinel system.
\newblock {\it Eur J Clin Pharmacol} \textbf{54(6)}, 483--488 (1998)

\bibitem{Bate1998}
Bate, A., Lindquist, M., Edwards, I.R., et~al: A bayesian neural network method
  for adverse drug reaction signal generation.
\newblock {\it Eur J Clin Pharmacol} \textbf{54}, 315--321 (1998)

\bibitem{MiniSentinel2008}
Behrman, R.E., Benner, J.S., Brown, J.S., et~al: Developing the sentinel system
  - a national resource for evidence development.
\newblock { \it N Engl J Med} \textbf{364}, 498--499 (2011)

\bibitem{Brown2007}
Brown, J.S., Kulldorff, M., Chan, A., et~al: Early detection of adverse drug
  events within population-based health networks: application of sequential
  testing methods.
\newblock {\it Pharmacoepidemiol Drug Saf} \textbf{16}, 1275--1284 (2007)

\bibitem{Coloma2011}
Coloma, P.M., Schuemie, M.J., Trifiro, G., et~al: Combining electronic
  healthcare databases in europe to allow for large-scale drug safety
  monitoring: the eu-adr project.
\newblock {\it Pharmacoepidemiol Drug Saf} \textbf{20}, 1--11 (2011)

\bibitem{Curtis2008}
Curtis, J.R., Cheng, H., Delzell, E., et~al: Adaptation of bayesian data mining
  algorithms to longitudinal claims data: coxib safety as an example.
\newblock {\it Med Care} \textbf{46}, 969--975 (2008)

\bibitem{DuMouchel1999}
DuMouchel, W.: Bayesian data mining in large frequency tables, with an
  application to the {FDA} spontaneous reporting systemt.
\newblock {\it Amer Statistician} \textbf{53}(3), 177--190 (1999)

\bibitem{evans2001}
Evans, S.J.W., Waller, P.C., Davis, S.: Use of proportional reporting ratios
  ({PRR}s) for signal generation from spontaneous adverse drug reaction
  reports.
\newblock Pharmacoepidemiol and Drug Saf \textbf{10(6)}, 483--486 (2001)

\bibitem{inps2011}
{INPS, A Cegedim Company}: Welcome to inps, http://www.inps4.co.uk/ (2011,
  accessed 25 jan 2012)

\bibitem{Jin2006}
Jin, H., Chen, J., Kelman, C., et~al: Mining unexpected associations for
  signalling potential adverse drug reactions from administrative health
  databases.
\newblock {\it PAKDD} pp. 867--876 (2006)

\bibitem{Jin2010}
Jin, H.W., Chen, J., He, H., et~al: Signaling potential adverse drug reactions
  from administrative health databases.
\newblock {\it IEEE Trans Knowl Data Eng} \textbf{22(6)}, 839--853 (2010)

\bibitem{BNF}
{Joint Formulary Committee . {\it British National Formulary}}: 62 ed.
\newblock London: BMJ Group and Pharmaceutical Press (2011)

\bibitem{Lewis2005}
Lewis, J.D., Bilker, W.B., Weinstein, R.B., Strom, B.L.: The relationship
  between time since registration and measured incidence rates in the general
  practice research database.
\newblock Pharmacoepidemiol Drug Saf \textbf{14(7)}, 443--451 (2005)

\bibitem{Noren2010}
Noren, G.N., Hopstadius, J., Bate, A., et~al: Temporal pattern discovery in
  longitudinal electronic patients records.
\newblock {\it Data Min Knowl Disc} \textbf{20}, 361--387 (2010)

\bibitem{Primohamed2004}
Piromohamed, M., James, S., Meakin, S., et~al: Adverse drug reactions as cause
  of admission to hospital: prospective analysis of 18820 patients.
\newblock {\it Br Med J} \textbf{329}, 15--19 (2004)

\bibitem{Puijenbroek2002}
van Puijenbroek, E.P., Bate, A., Leufkens, H.G.M., et~al: A comparison of
  measures of disproportionality for signal detection in spontaneous reporting
  systems for adverse drug reactions.
\newblock {\it Pharmacoepidemiol Drug Saf} \textbf{11}(1), 3--10 (2002)

\bibitem{Rosenberg2005}
Rosenberg, L., Coogan, P., Palmer, J.: Case-Control Surveillance, in
  {Pharmacoepidemiology}, fourth edition (ed b. l. strom) edn.
\newblock John Wiley \& Sons, Chichester, UK (2007)

\bibitem{Ryan2012}
Ryan, P.B., Madigan, D., Stang, P.E., Overhage, J.M., Racoosin, J.A.,
  G.Hartzema, A.: Empirical assessment of methods for risk identification in
  healthcare data: Results from the experiments of the observational medical
  outcomes partnership.
\newblock Statist. Med. \textbf{31}, 4401--4415 (2012)

\bibitem{Ryan2009}
Ryan, P.B., Powell, G., Pattishall, E., Beach, K.: Performance of screening
  multiple observational databases for active drug safety surveillance.
\newblock International Society of Pharmacoepidermiology:Providence,RI,USA,
  (2009)

\bibitem{Schuemie2012b}
Schuemie, M.J., Coloma, P.M., Straatman, H., Herings, R.M.C., Trifirò, G.,
  Matthews, J.N., Prieto-Merino, D., Molokhia, M., Gini, L.P.R., Innocent, F.,
  Mazzaglia, G., Picelli, G., Scotti, L., van~der Lei, J., Sturkenboom,
  M.C.J.M.: Using electronic health care records for drug safety signal
  detection: A comparative evaluation of statistical methods.
\newblock Medical Care. \textbf{50(10}, 890--897 (2012)

\bibitem{Shephard2011}
Shephard, E., Stapley, S., Hamilton, W.: The use of electronic databases in
  primary care research.
\newblock {\it Fam Pract} \textbf{28(4)}, 352--354 (2011).
\newblock \doi{10.1093/fampra/cmr039}

\bibitem{Story1974}
Story, N.L.: Sexual dysfunction resulting from drug side effects.
\newblock {\it J Sex Res} \textbf{10(2)}, 132--149 (1974)

\bibitem{ATC}
{WHO Collaborating Centre for Drug Statistics Methodology}: {ATC}
  classification index with {DDD}s, 2013.
\newblock {Oslo} (2012)

\bibitem{Wilke2011}
Wilke, R.A., Xu, H., Denny, J.C., et~al: The emerging role of electronic
  medical records in pharmacogenomics.
\newblock {\it Int J Clin Pharmacol Ther} \textbf{89(3)}, 379--386 (2001)

\bibitem{ICD10}
{World Health Organization}: International Statistical Classification of
  Diseases and Related Health Problems (The) {ICD}-10. 2010 Edition.
\newblock Nonserial Publications (2011)

\bibitem{Zhou2013}
Zhou, X., Murugesan, S., Bhullar, H., Liu, Q., Cai, B., Wentworth, C., Bate,
  A.: An evaluation of the thin database in the omop common data model for
  active drug safety surveillance.
\newblock Drug Saf \textbf{36}, 119--134 (2013)

\bibitem{Zorych2011}
Zorych, I., Madigan, D., Ryan, P., et~al: Disproportionality methods for
  pharmacovigilance in longitudinal observational databases.
\newblock {\it Stat Methods Med Res} \textbf{0(0)}, 1--18 (2011)

\end{thebibliography}

\end{document}